\documentclass[letterpaper]{article} 
\usepackage{aaai25}  
\usepackage{times}  
\usepackage{helvet}  
\usepackage{courier}  
\usepackage[hyphens]{url}  
\usepackage{graphicx} 
\usepackage{amsmath} 
\usepackage{natbib}  
\usepackage{caption} 
\usepackage[table,xcdraw]{xcolor}
\usepackage{algorithm}
\usepackage{algorithmic}

\usepackage{newfloat}
\usepackage{listings}
\DeclareCaptionStyle{ruled}{labelfont=normalfont,labelsep=colon,strut=off} 
\floatstyle{ruled}
\newfloat{listing}{tb}{lst}{}
\floatname{listing}{Listing}
\title{Zero-Shot Learning in Industrial Scenarios: New Large-Scale Benchmark, Challenges and Baseline}
\author{Zekai Zhang\textsuperscript{1}, Qinghui Chen\textsuperscript{1}, Maomao Xiong\textsuperscript{1}, Shijiao Ding\textsuperscript{1}, Zhanzhi Su\textsuperscript{2}, Xinjie Yao\textsuperscript{3}, Yiming Sun\textsuperscript{4}, Cong Bai\textsuperscript{5}, Jinglin Zhang\textsuperscript{\rm 1}\thanks{Corresponding author.}
}
\affiliations{
    \textsuperscript{\rm 1}School of Control Science and Engineering, Shandong University, Jinan, China\\
    \textsuperscript{\rm 2} National Supercomputing Center in Jinan, Qilu University of Technology, Jinan, China\\
    \textsuperscript{\rm 3} College of Intelligence and Computing, Tianjin University, Tianjin, China\\
    \textsuperscript{\rm 4} School of Automation, Southeast University, Nanjing, China\\
    \textsuperscript{\rm 5} College of Computer Science and Technology, Zhejiang University of Technology, Hangzhou, China\\
    \{202420810, 202420785, 202314866, 202234879\}@mail.sdu.edu.cn, 10431220602@stu.qlu.edu.cn, yaoxinjie@tju.edu.cn, sunyiming@seu.edu.cn, congbai@zjut.edu.cn, jinglin.zhang@sdu.edu.cn
}

\usepackage{bibentry}

\begin{document}

\maketitle

\begin{abstract}
Large Visual Language Models (LVLMs) have achieved remarkable success in vision tasks. However, the significant differences between industrial and natural scenes make applying LVLMs challenging. Existing LVLMs rely on user-provided prompts to segment objects. This often leads to suboptimal performance due to the inclusion of irrelevant pixels. In addition, the scarcity of data also makes the application of LVLMs in industrial scenarios remain unexplored. To fill this gap, this paper proposes an open industrial dataset and a Refined Text-Visual Prompt (RTVP) for zero-shot industrial defect detection. First, this paper constructs the Multi-Modal Industrial Open Dataset (MMIO) containing 80K+ samples. MMIO contains diverse industrial categories, including 6 super categories and 18 subcategories. MMIO is the first large-scale multi-scenes pre-training dataset for industrial zero-shot learning, and provides valuable training data for open models in future industrial scenarios. Based on MMIO, this paper provides a RTVP specifically for industrial zero-shot tasks. RTVP has two significant advantages: First, this paper designs an expert-guided large model domain adaptation mechanism and designs an industrial zero-shot method based on Mobile-SAM, which enhances the generalization ability of large models in industrial scenarios. Second, RTVP automatically generates visual prompts directly from images and considers text-visual prompt interactions ignored by previous LVLM, improving visual and textual content understanding. RTVP achieves SOTA with 42.2\% and 24.7\% AP in zero-shot and closed scenes of MMIO.
\end{abstract}

\section{Introduction}
Product defect detection tasks in industrial scenarios play an important role in ensuring the safety of users. Expert models \cite{wang2022yolov7,2021YOLOX,li2022yolov6} in industrial scenarios usually use single-modal data from a single field and strictly follow class-aware methods, which limits the ability of model to process multi-scene data and generalize to open datasets. Recently, the development of Segment Anything series (SAMs) \cite{sam,fastsam,fastersam} has shown powerful interactive and strong zero-shot capabilities in remote sensing, medicine, and other fields. The uniqueness of SAM lies in the design of human-computer interactive prompts, which allows segmentation based on user-provided point, line, and box prompts. \cite{1,2,3,4,5,6,7,8,9,10,11,12,13,zhang2026novel,zhang2023idd,zhang2024representation,zhang2026unification}

\begin{figure}[!t]
\centering
\includegraphics[width=3in]{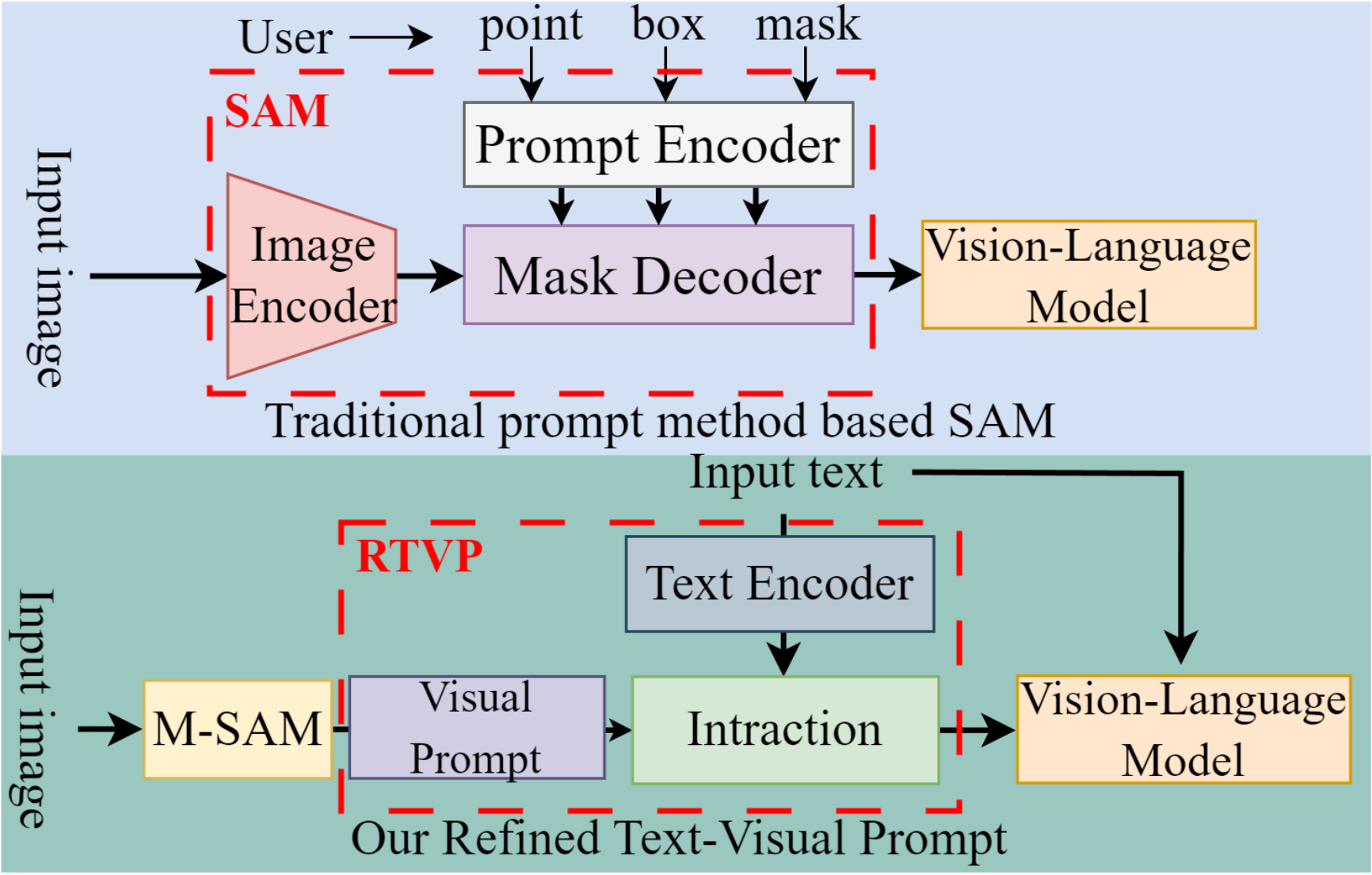}
\caption{Comparison between traditional prompting methods and RTVP. RTVP solves the subjectivity of traditional manual prompts and introduces text to further refine the semantics.}
\label{Fig1}
\end{figure}

However, since the environment, scale and distribution of images in industrial scenes are significantly different from natural scenes (\textbf{Figure \ref{Fig2}-a}), there are many significant challenges in applying SAM's pre-training-prompt paradigm in industrial scenes. As shown in \textbf{Figure \ref{Fig1}}, the existing SAM prompt mode needs to rely on the user's prompts (point, box, mask) \cite{sam} to segment objects when processing complex scenes. The user's familiarity will significantly affect specific prompts' effect and introduce irrelevant or noisy pixels.

To address the above problems, some methods \cite{matcher,personalize} combine semantic models \cite{resnext,resnet} to obtain pseudo masks of objects. CPT \cite{yao2024cpt} and ReCLIP \cite{subramanian2022reclip} use visual prompt to establish relationships between instances. On above basis, Hu et al. \cite{hu2023efficiently} designed a sampling strategy to extract points and bounding boxes from the pseudo template as prompts for SAM to segment the object image. CoCoOp \cite{yan2023cocoopter} turns the image-generated prompt into a conditional input and dynamically combines it with the language prompt. These methods ignore false positives in pseudo masks and rely hyperparameter sensitivity. Therefore, it heavily depends on the quality of pseudo masks and has poor generalization ability. To address the above problems, this paper proposes a Refined Text-visual prompt (RTVP), which improves the zero-shot capability of VLM in industrial scenarios. Based on the zero-shot of Mobile-SAM \cite{fastersam} in natural scenes, RTVP further enhances its generalization ability in industrial scenes. As shown in \textbf{Figure \ref{Fig1}}, considering the subjectivity and noise of user prompts, RTVP introduces expert-guided mechanism based on Mobile-SAM to generate coarse-grained segmentation features and encode the segmentation features into a low-dimensional space. Following the unique sparse nature of industrial images, this paper perform spatial-channel pixel activation on the segmentation features, extract uncertainty score of the object. Then, this paper introduces a sparse modelling sample selection strategy to extract the semantic clues from the enhanced features through the uncertainty score and obtain a refined Visual Prompt. Finally, the refined Visual Prompt is interacted with the text prompt to generate prompt embeddings of semantically specific objects. Based on the inherent capabilities of Mobile-SAM, RTVP enhances the model's visual-language understanding and generalization capabilities, especially in open-world and cross-domain scenarios.

Another challenge to applying SAM in industrial scenarios is that the existing data in industrial inspection is single, and it is impossible to find a unified multi-domain generalized industrial scene dataset. To solve the above problem, this paper creates a large-scale Multi-Modal Industrial Open dataset called MMIO-80K. MMIO consists of more than 80K+ samples converted from 18 different industrial defect datasets, including various product defects in major industrial categories such as metallurgy, automobile manufacturing, precision electronics, textiles, daily necessities, and wood processing. MMIO is tailored for the unique feature distribution in industrial zero-shot detection, effectively alleviating the lack of industrial domain expertise of LVLMs. To the best of our knowledge, MMIO-80K is the first open dataset proposed for industrial zero-shot detection, and MMIO-80K can catalyze the development of LVLMs in industrial openness.

In the detection task of MMIO closed scenes, RTVP significantly improved mAP compared to YOLOv8, YOLOv9 and other field expert models. In the zero-shot detection task in MMIO, RTVP surpasses most benchmark models. Generalization experiments on COCO and LVIS datasets also show that RTVP can generalize detection in natural scenes. In summary, the core contributions of this paper are as follows: \footnote[1]{Extended version: https://github.com/hellozzk/MMIO}

$\bullet$\textbf{MMIO-80K:} To the best of our knowledge, this paper constructed the first object detection data set MMIO-80K for industrial open scenarios. MMIO-80K consists of more than 80K samples, effectively alleviating the lack of domain expertise in industrial open scenarios.

$\bullet$\textbf{RTVP:} This paper proposes refined learnable text-visual prompts to improve the zero-shot detection capability of visual-language models in industrial scenarios. RTVP does not require users to provide specific prompts for each image, which reduces the user's usage burden and noise and effectively improves the knowledge and understanding ability of LVLMs in industrial domains.

$\bullet$\textbf{Superior performance:} Extensive experiments show that RTVP has superior performance in industrial open scene detection. Therefore, RTVP represents a significant advancement in LVLMs for industrial detection, providing a general method for mutual visual language understanding in industrial scenarios.

\section{Related Work}
\subsection{Application of Vision-Language Models}
In recent years, pre-trained Large language models, such as GPT-4 \cite{gpt4}, Llava \cite{llava}, etc., have shown strong zero-shot learning capabilities in natural language processing. Subsequently, pre-trained visual-language models such as CLIP \cite{clip}, BLIP-2 \cite{li2023blip}, etc., have been extended to computer vision. Currently, there are two methods to apply large pre-trained models. One is to use the segmentation results of large pre-trained models as prior information to assist downstream tasks, which requires additional intermediate layer fine-tuning of model. For example, Ahmadi et al. \cite{ahmadi2023application} used the segmentation results of SAM as prior information in crack and other defect detection. Wu et al. \cite{wu2023medicalsam} inserted the Adapter module into SAM for medical image segmentation tasks. Another method uses prompts to guide the pre-trained model transfer to the object domain. For example, Xu et al. \cite{xu2023eviprompt} proposed an untrained evidence prompt generation method, incorporating human prior information into prompts. Zhang et al. \cite{jie2023adaptershadow} proposed generation network for the shadow detection to generate dense point prompts. The above two methods easily consume computing resources and cannot guarantee the training effect of the domain layer. In addition, the visual prompts are not refined and do not consider the importance of text prompts. In contrast, in terms of generating prior information, this paper introduces an expert model to assist Mobile-SAM. In terms of prompts, this paper uses refined text-visual prompts to provide richer object semantic information. 
\subsection{Prompt for Zero-shot Learning}
Prompt technology originated from NLP. The prompt was subsequently used to guide zero-shot learning of large pre-trained models. However, prompts often rely on artificial features, leading to user burden and noise introduction. Recently, automated prompt training methods have been widely used in zero-shot tasks. For example, AutoPrompt \cite{shin2020autoprompt} used gradient-based methods to generate prompt templates automatically. With the development of large models, fine-grained visual prompts are widely used in zero-shot detection scenarios. For example, CPT \cite{yao2024cpt} introduces coloured object boxes as markers on the image. However, the prompts contain a lot of noise. To solve this problem, FGVP \cite{yang2024fine}, VRP-SAM \cite{sun2024vrp} etc., proposed refined visual prompts. However, these methods rely on the adaptability of SAM and cannot iteratively optimize the quality of prompts. In addition, these methods ignore the role of text prompts in zero-shot tasks. In contrast, this paper expresses more refined semantically specific object features through continuous optimized text-visual interaction prompts.

\begin{table*}[t]
\fontsize{10}{12}\selectfont\rmfamily
\centering

\begin{tabular}{ccccc}
\rowcolor[HTML]{FFFFFF} 
\hline
Dataset  & Classes & Number & Scene Category & Modal      \\
\hline
\rowcolor[HTML]{FFFFFF} 
MVTEC AD\cite{2020MVTec}  & 15      & 5,354   & 15             & Image      \\
\hline
\rowcolor[HTML]{FFFFFF} 
BTAD\cite{2021VT}& 3       & 2,830   & 3              & Image      \\
\hline
\rowcolor[HTML]{FFFFFF} 
VisA\cite{2020Deep}     & 12      & 10,821  & 12             & Image      \\
\hline
\rowcolor[HTML]{FFFFFF} 
Ind\cite{0Pixel}      & 30      & 600,000 & 11             & Image      \\
\hline
\rowcolor[HTML]{FFFFFF} 
MS-COCO\cite{COCO}  & 80      & 118,000 & 1             & Image      \\
\hline
\rowcolor[HTML]{FFFFFF} 
VOC2007\cite{2015voc}  & 20      & 9,963   & 1             & Image      \\
\hline
\rowcolor[HTML]{E6E0E0} 
MMIO-80K & 100     & 21,836  & 18             & Text,Image \\
\hline
\end{tabular}

\caption{Comparison of MMIO with zero-shot industrial defect dataset and general dataset.}
\label{Tab1}
\end{table*}

\section{Multi-Modal Industrial-Open Dataset}
\subsection{Dataset construction}
\begin{figure}[!t]
\centering
\includegraphics[width=3in]{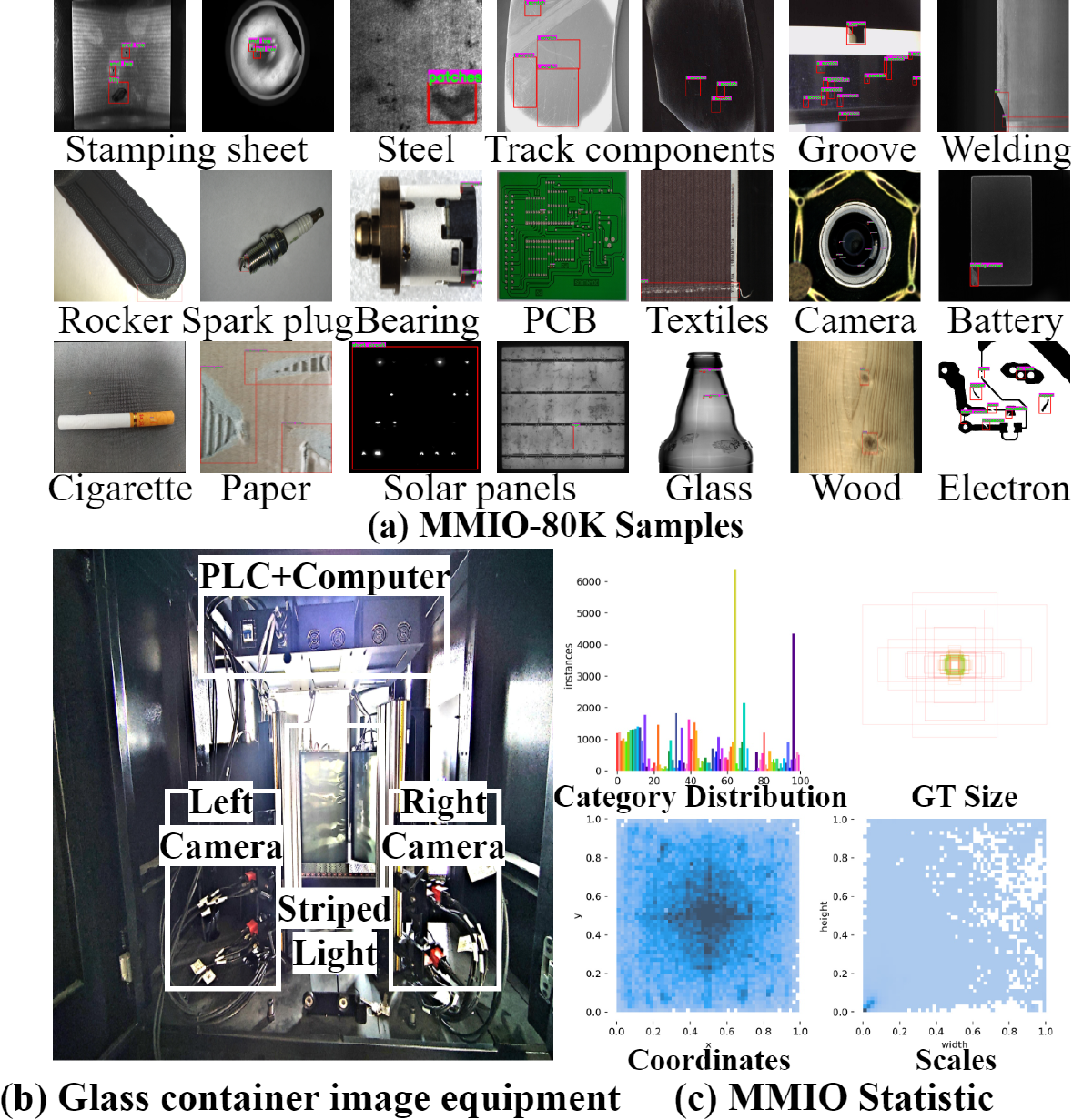}
\caption{MMIO statistical analysis. (a) MMIO example, which includes more than 18 types of industrial scenes. (b) The proposed Glass container image equipment. (c) MMIO data statistics results.}
\label{Fig2}
\end{figure}

This paper creates a large-scale Multi-Modal industrial open dataset named MMIO-80K. MMIO consists of more than 80K samples converted from 18 different major industrial scenes. Among them, the glass container dataset related to daily necessities is a subset collected by this paper. \textbf{Figure \ref{Fig2}-b} shows the glass container image acquisition equipment. This paper designs a multi-camera collaboration mechanism for cylindrical transparent objects such as glass container. Specifically, the conveyor belt conveys the container to the photoelectric gate to activating the three cameras on the left to take pictures. Then, the container will rotate 180 degrees, and the three cameras on the right will repeat the above process. A total of 625 pictures were collected for object detection and annotation. Each image has a resolution of 700 $\times$ 820 and has five categories (`oil', `black spot', `bubble', `plastering thread', and `quenched grain').

This paper obtains images and labels for the remaining 17 industrial scenarios through various enterprise open-source data. This paper delivers over 22,000 defect images initially summarized to professional technicians for quality screening, eliminates duplicate and blank images, and obtains 21,836 defect images in different fields. This paper re-adds attributes to each image based on experts' advice. This paper divides the defect into 6 super categories according to the main scenarios of intelligent manufacturing, and each super category contains multiple scene data. Since MMIO contains 80K+ defect entities, it is difficult to manually correspond the real category to the defect. Therefore, this paper uses CLIP to extract the semantic vector of each category and the cosine distance to calculate the similarity score between the semantic vector and the image-related semantics. In addition, the text-image matching pairs with the highest similarity are filtered by a threshold of 0.8. Finally, experts check the effect of text-image matching linked to 100 categories. We modified some similar objects. The number of samples of the same industrial product may be relatively small, and there may be similar defects: such as holes and irregular holes, but they still have similar categories. We divide them into seen and unseen categories to distinguish different categories of the same industrial product. In the end, MMIO has 21,836 images and 100 categories. 

\subsection{Dataset characteristics, statistics, and challenges}
\textbf{Features and statistics: }MMIO has 6 super categories and 100 attributes. \textbf{Figure \ref{Fig2}-c} summarizes the distribution of MMIO categories and instance sizes. Different product types and manufacturing in various industrial fields will produce different defects, among which small-sized defects account for the vast majority. In particular, MMIO stands out with rich attribute annotations, covering a wide range of standard industrial manufacturing categories, making it particularly suitable for the complex task of industrial zero-shot detection. The 100 classes in MMIO can be divided into 6 super categories: metallurgy, automobile manufacturing, precision electronics, textiles, daily necessities, and wood processing. Examples of each category are shown in \textbf{Figure \ref{Fig2}-a}. \textbf{Table \ref{Tab1}} compares general zero-shot industrial defect comprehensive datasets and general datasets, which indicates that MMIO has significant advantages in the number of scenes and semantic annotations. To our knowledge, MMIO is the first multi-scene defect dataset for industrial zero-shot detection. Unlike only nature scene general object detection datasets such as MS-CCOCO \cite{COCO} and VOC2007 \cite{2015voc}, MMIO specializes in text annotation enhancement for zero-shot learning tasks, providing a more challenging and relevant benchmark for zero-shot learning.

\textbf{Dataset processing:} MMIO is divided into a visible class training dataset and an invisible class test set for zero-shot tasks. Among them, the visible class contains 18,811 images and annotations, and the invisible class contains 3025 images and annotations. This paper also provides a training test set for closed scenario tasks, with a test and training set ratio of 20\% and 80\%.

\begin{figure*}[!h]
\centering
\includegraphics[width=7in]{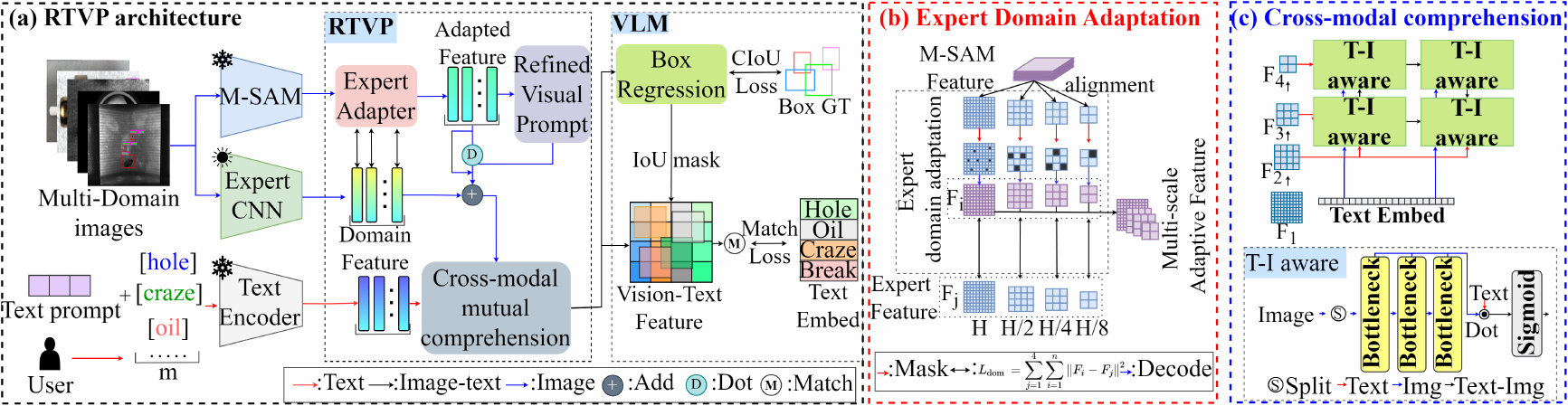}
\caption{TRVP framework. (a) The specific architecture of the RTVP. (b) This paper constructs a CNN expert model to assist Mobile-SAM in rapid domain adaptation. Refined Visual Prompt establishes a sample selection strategy to generate refined visual prompts. (c) The interaction between text and refined visual prompts deepens VLM's semantic understanding of relevant areas of the image.}
\label{Fig3}
\end{figure*}

\textbf{Dataset challenges:} MMIO faces material changes, imaging limitations, and internal-external interference. First, the product materials vary greatly, resulting in drastic scale changes in MMIO. The defect detection model needs to perceive object areas of different scales sensitively. Secondly, environmental interference and intra-class correlation make the discriminant information of defects unclear, which requires the model to be robust. The third is the long-tail distribution problem. Industrial scenarios need to seek balanced detection effects in extremely unbalanced category distributions.
\section{Proposed Method}

\subsection{Problem Definition}

The goal of the zero-shot framework is to recognize objects that have never appeared in the domain under pre-training with domain-specific text $Y\subseteq \varphi^{N} =\left\{t^{1},...,t^{N} \right\}$ and image $I\subseteq\delta ^{N} =\left\{ I^{1},...,I^{N}\right\}$. This paper provides a training dataset $D_{s}$ containing image-text pairs of $C_{s}$ visible categories. Let $z_{s}=\left \{1,..., C_{s}\right\}$ and $z_{u}=\left \{1,..., C_{u}\right\}$ are the label sets of visible and invisible categories, respectively. $z_{s}\cap z_{u}=\phi$. $D=D_{s}+D_{u}$ is the image-label space set of visible and invisible classes. Let the text set $Y=y_{s}+y_{u}$. During training, the model extracts the semantic information of $y_{s}$ contained $z_{s}$ and accurately matches $z_{s}$ to the relevant area of image $I$. A test set $D_{t}$ contains $D_{s}$ and $D_{u}$ in the zero-shot stage. The goal of the zero-shot task is to optimize a model from $D_{s}$ and detect the invisible category $C_{u}$ in $D_{u}$ through the user-defined invisible text prompt $y_{u}$ ($y_{u}$ contains the semantic information of $z_{u}$). For the visible category, this paper tests the $D_{s}$ category accuracy in $D_{t}$ to measure the effect of the visible class.

\subsection{Refined Text-Visual Prompt}

\textbf{Figure \ref{Fig3}} shows the detailed structure of the zero-shot framework proposed in this paper. Traditional zero-shot methods usually use RoIAlign \cite{2017Mask} or manually set points, lines, boxes and other prompts to obtain regional prompts. However, the coarse feature area leads to excessive noise, and the prompt cannot be iteratively optimized. The RTVP includes three important innovations: expert-assisted domain adaptation, refined visual prompter, and cross-modal prompt interaction, significantly improving the understanding ability of Mobile-SAM in industrial scenarios.

\textbf{Expert-assisted Domain Adaptation:} Although Mobile-SAM has strong zero-shot capabilities, obtaining refined visual prompt is still challenging. The challenge is that Mobile-SAM trained in natural scenes makes it difficult to transfer to industrial scenes effectively. Previous methods usually insert learnable layers into SAM to make it conform to the feature distribution of the target domain. However, above methods typically do not effectively supervise the adaptation layer, resulting in a poor migration effect. To solve the above problems, this paper proposes expert-assisted domain adaptation (\textbf{Figure \ref{Fig3}-b}). Expert-assisted domain adaptation uses an expert CNN architecture trained in industrial scenes to provide expert knowledge to Mobile-SAM. The principle is that the convergence speed of the pre-trained expert CNN in industrial scenes is faster than the domain adaptation layer of Mobile-SAM, so it can quickly give Mobile-SAM adequate supervision to achieve domain transfer. The expert model is not frozen during training because the iteratively optimized expert CNN can better provide domain expertise.

In practice, given a CNN-based domain expert model $M_{Expert}$, this paper first pre-trains it on visible categories in the training dataset $D_{s}$ to obtain the expert model multi-scale feature $F_{j} ,j\subseteq \left \{1,2,3,4\right\}$. To enhance the multi-scale information of industrial defects, this paper performs multi-scale sampling on the output feature of Mobile-SAM to obtain the multi-scale feature $F_{i} ,i\subseteq \left \{1,2,3,4\right\}$. As shown in \textbf{Figure \ref{Fig3}-b}, this paper masks and reconstructs the multi-scale features from Mobile-SAM to improve the robustness of domain adaptation. Unlike MAE \cite{he2022mae}, this paper reconstructs the domain expert information instead of the original input. This paper divides the features into $N$ groups according to the patch size to obtain the neighbourhood multi-scale feature $F_{i}^{N} ,(N\subseteq 2^{k} ,k\subseteq N^{+} )$. Then, this paper uniformly masks part of the patch according to the random sampling method, and the two-dimensional index of each patch is defined as $P_{i}^{c} \subseteq [(P_{1}^{x},P_{1}^{y}),...,(P_{N}^{x},P_{N}^{y})]$. Randomly sample $M$ of the indexes to obtain a binary mask and multiply it with $F_{i}$ to get the masked multi-scale feature $F_{i}^{M}$. The formula for the above process is as follows:
\begin{equation}
\begin{array}{c}
\operatorname{Mask}_{x, y}^{\mathrm{j}}=\left\{\begin{array}{c}
0,(x, y \in \mathrm{S}) \\
1, \text { Otherwise } 
\end{array}\right.,\\
F_{i}^{M}=P\left(F_{i}\right) \times \operatorname{Mask}_{x, y}^{\mathrm{j}},
\end{array}
\end{equation}
Among them, $s=P_{i}^{c}$ is the range of the extracted $M$ indexes. ${Mask}_{x, y}^{\mathrm{j}},j\subseteq \left \{1,2,3,4\right\}$ is a binary mask, and $P$ is the image segmentation operation. After obtaining the $F_{i}^{M}$, this paper builds a decoder to reconstruct the masked multi-scale features to get $F_{i}^{R}$. $F_{i}^{R}$ can be considered a coarse-grained visual prompt. The mask decoder consists of three layers of convolution and deconvolution. The formula is as follows:
\begin{equation}
\resizebox{0.9\hsize}{!}{$
F_{i}^{R}=\sum_{1}^{3} \operatorname{Relu}\left(\operatorname{Conv}\left(F_{i}^{M}\right)\right) \sum_{1}^{3} \operatorname{Relu}\left(\operatorname{Conv^{T}}\left(F_{i}^{M}\right)\right)$},
\end{equation}
To solve the problem of insufficient optimization of the existing methods, this paper introduces a multi-scale domain optimization function to optimize the domain adaptation layer. It is worth noting that this paper will stop optimizing the domain adaptation layer when the optimisation function is infinitely small. The optimization function is as follows:
\begin{equation}
f_{\text {optimization}}= \sum_{j=1}^{4} \sum_{i=1}^{4}\left\|F_{j}-F_{i}^{R}\right\|.
\end{equation}
\textbf{Refined Visual Prompt: }This paper uses Mobile-SAM guided by domain expert knowledge to generate coarse-grained visual prompts automatically and establishes a sparse modelling sample selection strategy to obtain more refined visual prompts. Refined Visual Prompt can more accurately highlight target instances, reduce background interference, and retain global knowledge. Specifically, based on the coarse-grained visual prompt $F_{i}^{R}$ obtained by expert-assisted domain adaptation, the sparse modelling sample selection strategy activates the coarse-grained visual prompt through a learnable scorer to obtain an uncertainty score ($M_{k}$) for each pixel. The formula is as follows:
\begin{equation}
M_{k}=\varphi_{\mathrm{i}}^{\mathrm{s}}\left(\varphi_{\mathrm{i}}^{\mathrm{c}}\left(F_{i}^{R}\right)\right),
\end{equation}
Among them, $\varphi_{\mathrm{i}}^{\mathrm{c}}$ represents channel activation and $\varphi_{\mathrm{i}}^{\mathrm{s}}$ represents spatial activation. $i$ represents different features. The formula of $\varphi_{\mathrm{i}}^{\mathrm{c}}$ and $\varphi_{\mathrm{i}}^{\mathrm{s}}$ are as follows:
\begin{equation}
\begin{array}{l}
\resizebox{0.95\hsize}{!}{$
\varphi_{\mathrm{i}}^{\mathrm{c}}=\left(\sigma \mathrm{Q}_{\mathrm{k}}\left(\mathrm{W}_{\mathrm{j}}\left(\mathrm{X}_{\mathrm{C}}^{\operatorname{mean}}\left(F_{i}^{R}\right)\right)\right)+\sigma \mathrm{Q}_{\mathrm{k}}\left(\mathrm{W}_{\mathrm{j}}\left(\mathrm{X}_{\mathrm{C}}^{\max }\left(F_{i}^{R}\right)\right)\right)\right.$}, \\
\varphi_{\mathrm{i}}^{\mathrm{s}}=\sigma\left(\operatorname{Cat}\left[\mathrm{X}_{\mathrm{s}}^{\operatorname{mean}}\left(F_{i}^{R}\right) ; \mathrm{X}_{\mathrm{s}}^{\max }\left(F_{i}^{R}\right)\right]\right),
\end{array}
\end{equation}
Among them, $\sigma$ represents the activation function sigmoid, $\mathrm{Q}_{\mathrm{k}} \in\left[\mathrm{Q}_{1}, \mathrm{Q}_{2}\right], \mathrm{W}_{\mathrm{j}} \in\left[\mathrm{W}_{1}, \mathrm{~W}_{2}\right]$ belong to the complementary multi-layer activation perception weights,$X_{c}^{mean}$ represents the channel direction average, $X_{c}^{max}$ represents the channel direction maximum, $X_{s}^{mean}$ represents the spatial direction average, $X_{c}^{max}$ represents the spatial direction maximum, and $Cat$ represents channel superposition. The uncertainty score reflects the area with the most information. Traditional uncertainty methods usually select the samples with the highest scores. However, samples with significant uncertainty scores are not the optimal feature. Therefore, this paper creates a sparse sampling mechanism. Specifically, industrial images have significant sparse characteristics, and the sparse sampling mechanism selects the most high-frequency areas in the image to reduce the redundancy of irrelevant features. This paper uses patch segmentation with different neighbourhood sizes to retrieve the uncertainty score $M_{k}$. Specifically, given a set of segment patches $N_{h,w}^{p}$, a sparse selection mechanism is used to select high-frequency pixels. This paper performs pixel-level mean on the feature space inside the $N_{h,w}^{p}$. Then, completes sparse sample selection by selecting the top-k pixels, which pixel values inside the patch are more significant than the mean. The process can be described as the following formula:
\begin{equation}
\resizebox{0.9\hsize}{!}{$
\mathrm{M}_{\text {prompt}}^{\mathrm{i}, \mathrm{j}}=\left(\operatorname{Argmax}\left(\mathrm{N}_{h, \mathrm{w}}^{\mathrm{p}} . \text { mean, } \mathrm{D}_{\mathrm{i}, \mathrm{j}}^{\mathrm{c} s \mathrm{~s}}\right)\right),\left(\mathrm{i}, \mathrm{j} \in \mathrm{N}_{h, \mathrm{w}}^{\mathrm{p}}\right)$},
\end{equation}
Among them, $D_{i,j}^{c,s}$ is any pixel in $N_{h,w}^{p}$, $N_{h,w}^{p}.mean$ represents the pixel-level mean in the neighbourhood space, and $M_{prompt}^{i,j}$ is the selected sparse sample. After multiple selections, $M_{prompt}^{i,j}$ can be considered as a refined visual prompt. The sparse sampling mechanism can more accurately describe the contours and other details of the object. However, the lack of adequate supervision in the sparse selection mechanism causes a small number of selected feature points to deviate from the object. Therefore, this paper introduces CIoU-optimized feature activation to make the selected pixels fall within the ground truth. The optimization mechanism uses an additional detection head for regression prediction and continuously optimizes the visual prompt through CIoU.

\textbf{Cross-modal Interactive Visual-Text Prompt:} To further refine the semantic features related to the object, this paper interacts the text prompt with the visual prompt to refine the visual prompt in the multi-scale feature extraction. As shown in \textbf{Figure \ref{Fig3}-c}, given the text embedding $T_{i}$ from text encoder and the image feature $\mathrm{F}_{\mathrm{L}} \in \mathrm{R}^{\mathrm{C\times H\times W}}(\mathrm{L} \in\{1,2,3,4\})$. This paper adopts multi-scale image features and aggregates the text features into the image features using the maximum Sigmoid attention query text-image matching semantic features. The formula is as Eq-7.
\begin{equation}
\mathrm{F}_{\text {img-text }}=\mathrm{F}_{\mathrm{L}} \times Sigmoid\left(\operatorname{Argmax}\left(\mathrm{F}_{\mathrm{L}} \times \mathrm{T}_{\mathrm{i}}^{\mathrm{T}}\right)\right)^{\mathrm{T}}.
\end{equation}
\section{Experiments}
In this section, ablation studies and comparative experiments are performed to demonstrate the effectiveness of RTVP.
\subsection{Dataset and Evaluation Metrics}
This paper conducts experiments on MMIO-80K, MSCOCO, and LVIS datasets. For the MMIO zero-shot task, this paper split MMIO into 65 base classes for training and 35 novel classes for testing. For the MMIO closed task, this paper split MMIO into a training-test set ratio of 80\% and 20\%. To evaluate generalization, this paper performs closed-scene validation on COCO and zero-shot validation on LVIS. This paper uses the COCO and LVIS metrics to measure the model's accuracy.
\subsection{Implementation Details}
The experimental model is built on PyTorch 2.0.1, and the hardware environment is 4 Nvidia RTX 4090 GPUs. The model is trained 200 epochs using AdamW with 32 batches. The input image size is 640. The initial learning rate is 2e-3, the weight decay is 0.025, the text encoder (CLIP-Text Encoder) and Mobile-SAM-T are frozen during pre-training. The expert model is trained on MMIO seen classes for zero-shot task.

\subsection{Comparison with the State-of-the-art}
\textbf{MMIO closed scenario:} \textbf{Table \ref{Tab2}-a} compares general and expert defect models in MMIO closed scenarios. RTVP achieves SOTA in multi-scale AP and Recall, which indicates that RTVP is highly sensitive to multi-scale and long-tail distribution data sets. Compared with the traditional expert model's user training-detection mode, RTVP allows users to customize vocabulary and automatically generate refined prompts, providing accuracy that traditional expert models cannot provide. Compared with the recently proposed YOLOv10, AP of RTVP is improved by 0.7\%, and Recall is improved by 5.6\%. Experiments show that RTVP can also enhance the accuracy in closed scenarios. Beacuse of RTVP introduces expert model and industrial expert text labels to convey richer knowledge to VLM.

\begin{table*}[t]
\centering
\begin{tabular}{ccccc}
\hline
\multicolumn{5}{c}{\textbf{(a) MMIO Closed Scenarios}}       \\ 
\hline
\rowcolor[HTML]{FFFFFF} 
Method(general detection)           & AP $\uparrow$           & $\mathrm{AP}_{50}$ $\uparrow$   & Precision $\uparrow$    & Recall $\uparrow$       \\
\hline
\rowcolor[HTML]{FFFFFF} 
YOLOv8-S \cite{YOLOv8}        & 39.9          & 64.5          & 66.9          & 67.2          \\

\rowcolor[HTML]{FFFFFF} 
YOLOv8-M         & 40.7          & 67.5          & 71.4          & 64.4          \\

\rowcolor[HTML]{FFFFFF} 
YOLOv9 \cite{wang2024yolov9}         & 41.1          & 72.1          & 71.7          & 69.2          \\

\rowcolor[HTML]{FFFFFF} 
YOLOv10-S \cite{wang2024yolov10}       & 41.7          & 69.2          & 70.8          & 67.0          \\

\hline
Method(defect detection) & AP $\uparrow$ & AP$_{50}$ $\uparrow$ & Precision $\uparrow$ & Recall $\uparrow$ \\       \\
\hline
\rowcolor[HTML]{FFFFFF} 

\rowcolor[HTML]{FFFFFF} 
LiteYOLO-ID \cite{lite-yolo}       & 40.0          & 67.9          & 69.2          & 70.6          \\

\rowcolor[HTML]{FFFFFF} 
LF-YOLO \cite{lf}        & 41.1          & 72.6          & 66.9          & 70.6          \\
\rowcolor[HTML]{FFFFFF} 
Steel det \cite{9615164}       & 37.8          & 70.7          & 71.6          & 69.3          \\
\hline
\rowcolor[HTML]{E6E0E0} 
RTVP             & \textbf{42.4} & \textbf{76.1} & \textbf{74.8} & \textbf{72.6} \\
\hline
\multicolumn{5}{c}{\textbf{(b) MMIO Zero-shot}}                                           \\
\hline
\rowcolor[HTML]{FFFFFF} 
Method           & Params $\downarrow$       & AP $\uparrow$           & $\mathrm{AP}_{50}$ $\uparrow$   & $\mathrm{AP}_{75}$ $\uparrow$   \\
\hline
\rowcolor[HTML]{FFFFFF} 
MDETR \cite{kamath2021mdetr}         & 169M          & 18.1          & 17.7          & 20.3          \\

\rowcolor[HTML]{FFFFFF} 
GLIP-T \cite{glip}          & 232M          & 19.6          & 30.1          & 22.2          \\

\rowcolor[HTML]{FFFFFF} 
GLIPv2-T         & 232M          & 22.4          & 31.8          & 25.4          \\

\rowcolor[HTML]{FFFFFF} 
DetCLIP-T \cite{yao2022detclip}       & 155M          & 22.9          & 24.7          & 26.6          \\

\rowcolor[HTML]{FFFFFF} 
YOLO-W-S \cite{yoloworld}        & \textbf{77M}  & 21.2          & 32.1          & 24.6          \\

\rowcolor[HTML]{FFFFFF} 
Grounding DINO-T \cite{liu2023groundingdino}& 172M          & 20.2          & 29.4          & 23.9          \\
\hline
\rowcolor[HTML]{E6E0E0} 
RTVP             & 131M          & \textbf{24.7} & \textbf{35.7} & \textbf{27.3} \\
\hline

\end{tabular}

\caption{Comparison with expert models on MMIO closed scenarios and zero-shots.}
\label{Tab2}

\end{table*}

\textbf{MMIO Zero-shot:} \textbf{Table \ref{Tab2}-b} compares the common zero-shot methods. Compared with Grounding DINO-T, $AP$ and $AP_{50}$ are improved by 4.5\% and 6.3\%, respectively. For the latest YOLO-World-s, $AP$ and $AP_{50}$ are improved by 3.5\% and 3.6\%, respectively. Experiments show that RTVP can achieve the best performance with the litter parameters. This is because RTVP considers the sparse characteristics of industrial scenarios, which is helpful for industrial zero-shot tasks. RTVP uses Mobile-SAM and uncertainty sparse modelling to obtain refined visual prompt, and the interaction between text and visual prompt helps VLM achieve better zero-shot performance. It is worth noting that the $AP$ of all methods is very low, indicating that MMIO's tasks are more challenging and valuable and will promote subsequent industrial scenario zero-shot tasks.

\textbf{Comparison with COCO:} This paper verifies the generalization ability of RTVP in closed scenes on COCO \cite{COCO}. To achieve a fairer fully supervised comparison, this paper uses YOLOv8 as the baseline for retesting. The results of MS-COCO are shown in \textbf{Table \ref{Tab3}}. Compared with YOLOv8s, AP increased from 44.9\% to 47.2\%. RTVP achieved the best in $AP_{s}$ and $AP_{m}$, demonstrating that RTVP has the generalization ability to improve the accuracy of expert models in closed environments.

\begin{table*}
\centering
\begin{tabular}{ccccc}
\rowcolor[HTML]{FFFFFF} 
\hline
Method     & AP $\uparrow$           & $\mathrm{AP}_{s}$  $\uparrow$         & $\mathrm{AP}_{m}$ $\uparrow$          & $\mathrm{AP}_{l}$  $\uparrow$         \\
\hline
\rowcolor[HTML]{FFFFFF} 
PP-YOLOE-S \cite{ppyoloe} & 43.0          & 23.2          & 46.4          & 56.9          \\

\rowcolor[HTML]{FFFFFF} 
PP-YOLOE-M  & \textbf{49.0} & 28.6          & 52.9          & 63.8          \\

\rowcolor[HTML]{FFFFFF} 
YOLOv8-S \cite{YOLOv8}  & 44.9          & -             & -             & -             \\

\rowcolor[HTML]{FFFFFF} 
YOLOv9-S \cite{wang2024yolov9}  & 46.7          & 26.6          & 56.0          & 64.5          \\

\rowcolor[HTML]{FFFFFF} 
YOLOv9-M   & 51.1          & 33.6          & 57.0          & \textbf{68.0} \\

\rowcolor[HTML]{FFFFFF} 
YOLOv10-S \cite{wang2024yolov10} & 46.3          & -             & -             & -             \\
\hline
\rowcolor[HTML]{E6E0E0} 
RTVP       & 47.2          & \textbf{35.2} & \textbf{58.3} & 65.4 \\        
\hline
\end{tabular}
\caption{Comparative experiment on MS-COCO. This paper uses YOLOv8 as the baseline to test the RTVP.}
\label{Tab3}
\end{table*}

\textbf{Comparison with LVIS:} This paper evaluates the zero-shot generalization ability of RTVP on LVIS (\textbf{Table \ref{Tab4}}). Specifically, RTVP is pre-trained on the Object365 \cite{shao2019objects365} and GoldG datasets with YOLO-world-s as the baseline and fine-tuned on LVIS base. Compared with the baseline, RTVP improves by 0.6\% in AP with object365 pretraining, proving that RTVP's refined visual prompt improves VLM's zero-shot scene understanding ability. Compared with other models, RTVP's AP is still improved. With the increase of pre-training data, the performance of RTVP has improved, indicating that pre-training with a large amount of data improves accurate prompt expression.

\begin{table*}
\centering
\begin{tabular}{cccccc}
\rowcolor[HTML]{FFFFFF}
\hline
Method           & Pre-trained Data       & AP $\uparrow$           & $\mathrm{AP}_{r}$  $\uparrow$         & $\mathrm{AP}_{c}$ $\uparrow$          &$\mathrm{AP}_{f}$ $\uparrow$          \\
\hline
\rowcolor[HTML]{FFFFFF} 
GLIP-T \cite{glip}          & O365       & 17.8          & 13.5          & 12.8          & 22.2          \\
\rowcolor[HTML]{FFFFFF} 
YOLO-W \cite{yoloworld}      & O365       & 23.5          & 16.2          & 21.1          & 27.0          \\
\rowcolor[HTML]{FFFFFF} 
ViLD  \cite{gu2021open}           & O365       & -             & -             & 20.0          & \textbf{28.3} \\
\rowcolor[HTML]{E6E0E0} 
RTVP             & O365       & \textbf{24.1} & \textbf{17.3} & \textbf{22.4} & 27.9          \\
\hline
\rowcolor[HTML]{FFFFFF} 
Grounding-Dino-T \cite{liu2023groundingdino}& O365,GlodG & 25.6          & 14.4          & 19.6          & \textbf{32.2} \\
\rowcolor[HTML]{FFFFFF} 
GLIP-T \cite{glip}          & O365,GlodG & 24.9          & 17.7          & 19.5          & 31.0          \\
\rowcolor[HTML]{FFFFFF} 
YOLO-W-s \cite{yoloworld}         & O365,GlodG & 24.2          & 16.4          & 21.7          & 27.8          \\
\rowcolor[HTML]{E6E0E0} 
RTVP             & O365,GlodG & \textbf{26.8} & \textbf{19.5} & \textbf{23.4} & 30.7         \\
\hline
\end{tabular}
\caption{Zero-shot experiments of LVIS. This paper uses different datasets for pre-training.}
\label{Tab4}

\end{table*}

\subsection{Ablation Study}
\textbf{Component ablation study:} \textbf{Table \ref{Tab5}} shows the results of different component ablation study on MMIO. Expert domain adaptation can effectively improve the accuracy of zero-shot and closed scenes. Because the expert model can provide expert knowledge supervision to Mobile-SAM, it enhances the migration effect in the industrial field. The refined visual prompt is more conducive to zero-shot industrial detection. Because it uses sparse modelling of industrial images to help focus on key features. Cross-modal text-visual interaction is conducive to further refinement of semantic features related to objects, which is benefit to VLM identifying invisible class features accurately.

\begin{table}
\fontsize{10}{12}\selectfont\rmfamily
\centering
\begin{tabular}{
>{\columncolor[HTML]{FFFFFF}}c l
>{\columncolor[HTML]{FFFFFF}}c 
>{\columncolor[HTML]{FFFFFF}}c 
>{\columncolor[HTML]{FFFFFF}}c }
\hline
Method(Closed)               &                         & AP $\uparrow$                          & $\mathrm{AP}_{50}$  $\uparrow$                       &$\mathrm{AP}_{75}$ $\uparrow$                   \\
\hline
-Expert Domain Adapter        &                          & 30.2                         & 61.8                         & 29.1                         \\
-Refined Visual Prompt        &                          & 38.6                         & 72.9                         & 35.4                         \\
-Text-Visual Interation       &                          & 40.0                         & 73.6                         & 36.8                         \\
\cellcolor[HTML]{E6E0E0}RTVP & \cellcolor[HTML]{E6E0E0} & \cellcolor[HTML]{E6E0E0}42.4 & \cellcolor[HTML]{E6E0E0}76.1 & \cellcolor[HTML]{E6E0E0}38.2 \\
\hline
Method(Zero-shot)            &                          & AP $\uparrow$                          & $\mathrm{AP}_{50}$ $\uparrow$                        &$\mathrm{AP}_{75}$  $\uparrow$                       \\
\hline
-Expert Domain Adapter        &                          & 18.5                         & 27.4                         & 18.8                         \\
-Refined Visual Prompt        &                          & 20.0                         & 33.7                         & 23.9                         \\
-Text-Visual Interation       &                          & 22.8                         & 31.6                         & 25.1                         \\
\cellcolor[HTML]{E6E0E0}RTVP & \cellcolor[HTML]{E6E0E0} & \cellcolor[HTML]{E6E0E0}24.7 & \cellcolor[HTML]{E6E0E0}35.7 & \cellcolor[HTML]{E6E0E0}27.3 \\
\hline
\end{tabular}
\caption{Ablation study on MMIO closed scenes and zero-shots.}
\label{Tab5}

\end{table}

\begin{figure*}[!t]
\centering
\includegraphics[width=6in]{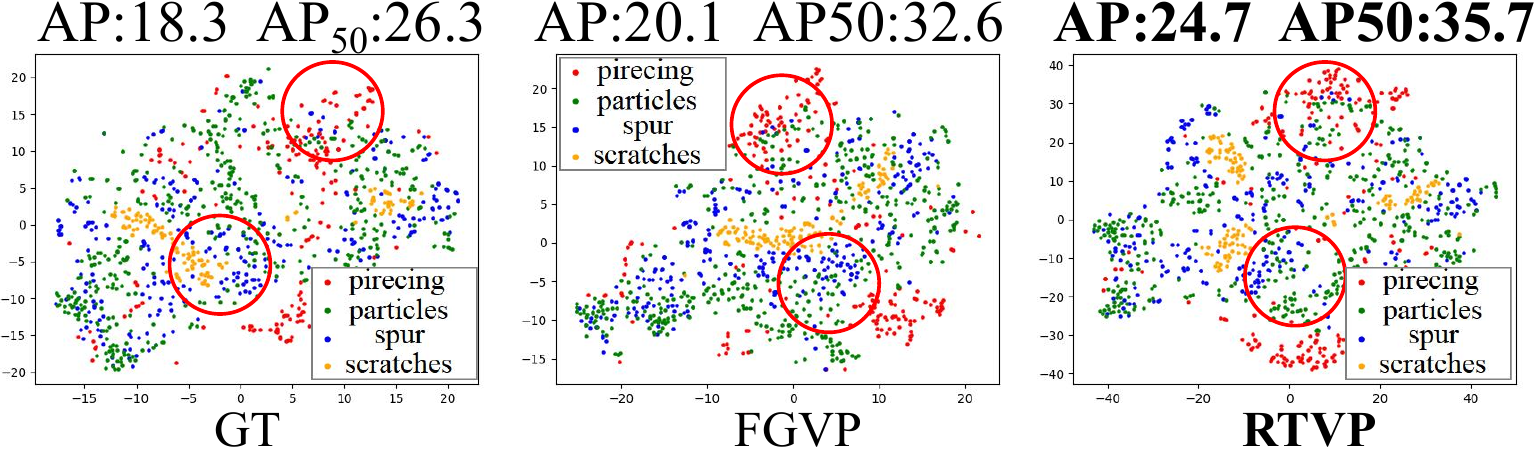}
\caption{t-SNE visualization in the MMIO zero-shot scenario. Compare with Ground Truth and FGVP, RTVP can generate most compact feature representations.}
\label{Fig4}
\end{figure*}

\textbf{Different prompts:} This paper tests the zero-shot detection effect of the ground truth as the visual prompt and FGVP (Fine-Grained Visual Prompting) on zero-shot task. Among them, FGVP \cite{yang2024fine} is reproduced with YOLO-World as the baseline. As shown in \textbf{Figure \ref{Fig4}}, this paper uses t-SNE \cite{t-sne} to visualize the image features of four industrial invisible categories on MMIO and the output features before the detector. Compared with using the ground truth and FGVP (only visual prompt), the features of RTVP show clear clusters. Experiments demonstrate the importance of refined visual prompt and text prompt interactions. As shown in AP, RTVP has the highest AP on MMIO, indicating that RTVP produces more obvious features, generates well-separated clusters for different classes, and promotes the learning of invisible classes.

\textbf{Different patch sizes:} The sparse modelling mechanism uses patches of different sizes to select high-frequency features. As shown in \textbf{Table \ref{Tab6}}, the patch size significantly affects the zero-shot task's accuracy. This is because the sparsity of industrial defects leads to obvious high-frequency features of defects. The smaller the patch, the more conducive it is to narrow the retrieval range, and it is easier to find high-frequency features.

\subsection{Qualitative Results}
\begin{figure}[!t]
\centering
\includegraphics[width=3in]{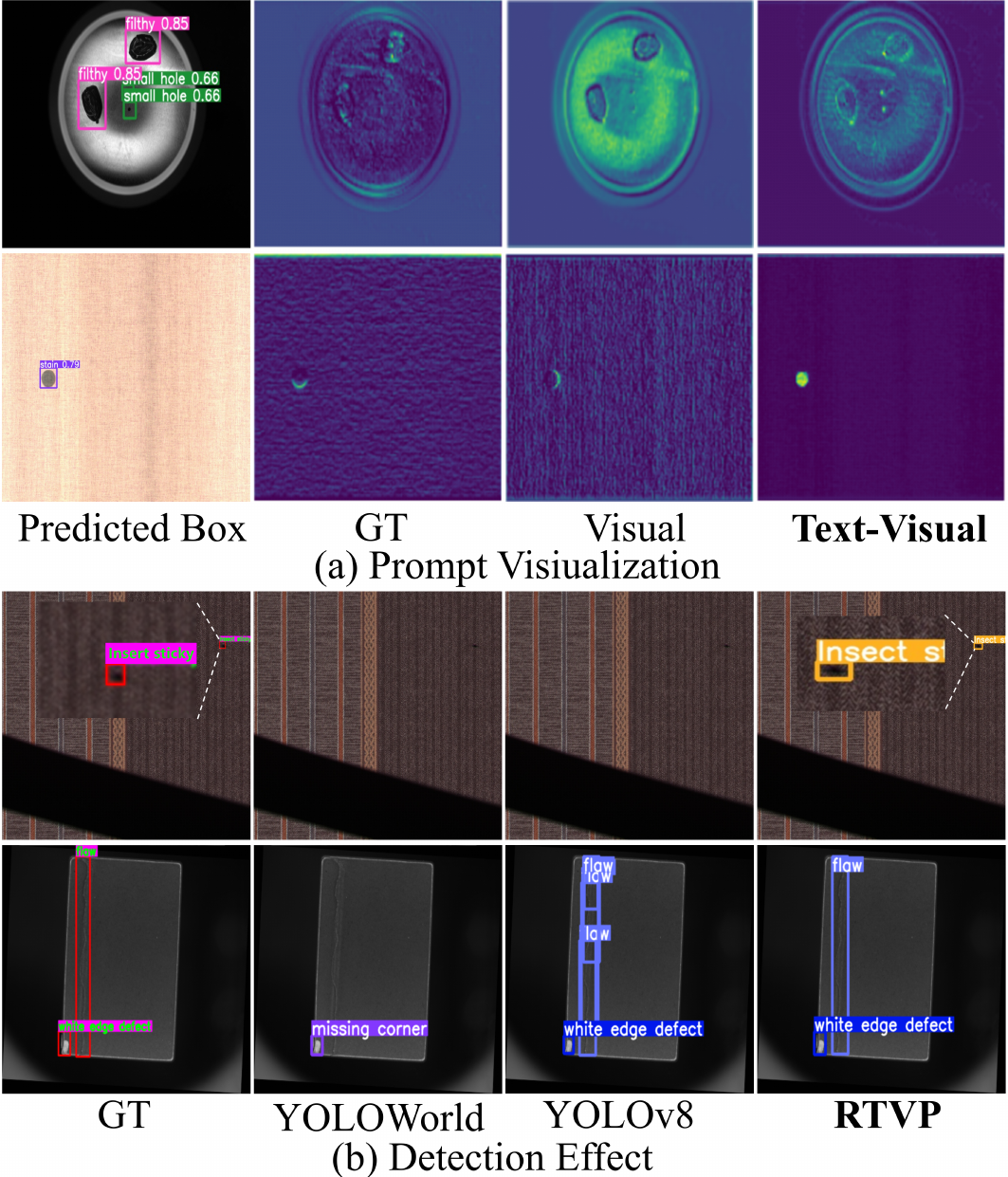}
\caption{RTVP Visualization. (a) The effect of different prompts on the feature space of VLM. The yellow area has high attention. (b) Detection effect.}
\label{Fig5}
\end{figure}

\textbf{Prompt visualization:} This paper maps and visualizes the VLM feature space after prompting. \textbf{Figure \ref{Fig5}-a} shows the impact of different prompts on the VLM feature space. Directly using the ground truth will introduce noise, which can make it difficult for VLM to understand the key areas of defects. The Visual Prompt feature saliency map shows that RTVP can significantly activate the high-frequency areas of defects, proving the effectiveness of the sparse modelling mechanism. The feature saliency map of the Text-Visual Prompt shows that introducing text features can further refine the semantic information related to the defect.

\begin{table}
\fontsize{10}{12}\selectfont\rmfamily
\centering
\begin{tabular}{cccc}
\rowcolor[HTML]{FFFFFF} 
\hline
Patches Size & AP $\uparrow$  & $\mathrm{AP}_{50}$ $\uparrow$ & $\mathrm{AP}_{75}$ $\uparrow$\\
\hline
\rowcolor[HTML]{E6E0E0} 
32           & 24.7 & 35.7 & 27.3 \\
\hline
\rowcolor[HTML]{FFFFFF} 
64           & 21.3 & 34.1 & 22.5 \\
\hline
\rowcolor[HTML]{FFFFFF} 
128          & 19.9 & 33.0 & 21.4 \\
\hline
\end{tabular}
\caption{Comparison of different patch sizes in MMIO zero-shot scenario.}

\label{Tab6}

\end{table}

\textbf{Detection result visualization:} \textbf{Figure \ref{Fig5}-b} visualizes the detection results of YOLOv8s, YOLOWorlds, and RTVP. The visualization results show that the other methods are prone to missed and imprecise detection. In contrast, RTVP effectively avoids the above problems.

\section{Conclusion}
The application of LVLMs in the industrial field is challenging due to the significant domain differences. To address the lack of professional data, this paper constructs the Multi-Modal Industrial Open Dataset (MMIO). MMIO contains diverse product defect data from major industrial categories, including 6 super categories and 18 subcategories. Based on MMIO, this paper provides Refined Text-Visual prompt (RTVP) for zero-shot tasks in industrial scenarios. Using expert-guided domain transfer, RTVP enhances the generalization ability of Mobile-Segment Anything in industrial scenarios. Secondly, RTVP proposes a text-visual interaction method to promote cross-modal mutual matching and understanding. Experiments have demonstrated the effectiveness of RTVP.

\section{Acknowledgments}
This work was supported in part by the Key Research and Development Program of Shandong Province of China under Grant 2023CXGC010112, in part by the Joint Fund for Regional Innovation Development of the National Natural Science Foundation of China (Grant U24A20221), in part by the National Key Research and Development Program of China under Grant 2022YFB4500602, in part by the Distinguished Young Scholar of Shandong Province under Grant ZR2023JQ025, in part by the Taishan Scholars Program under Grant tsqn202211290, and in part by the Major Basic Research Projects of Shandong Province under Grant ZR2022ZD32.

\bibliography{aaai25}

\end{document}